\pdfoutput=1

\documentclass[11pt]{article}

\usepackage{acl}

\usepackage{times}
\usepackage{latexsym}

\usepackage[T1]{fontenc}

\usepackage[utf8]{inputenc}

\usepackage{microtype}

\usepackage{inconsolata}

\usepackage{graphicx}

\usepackage{amsmath}
\usepackage{amssymb}
\usepackage{xcolor}
\usepackage{booktabs}
\usepackage{multirow}
\usepackage{subcaption}
\usepackage{colortbl}
\usepackage{xcolor}
\usepackage{array} 
\usepackage{soul}    
\usepackage{enumitem}
\definecolor{darkgreen}{RGB}{0, 100, 0}

\definecolor{lightblue}{rgb}{0.6, 0.8, 1} 
\definecolor{darkblue}{rgb}{0, 0, 0.6}    

\usepackage{amsmath}
\usepackage{amssymb}

\usepackage{algorithm}
\usepackage{algorithmicx}
\usepackage{algpseudocode}

\usepackage{microtype}
\usepackage{booktabs}

\title{ICR Probe: Tracking Hidden State Dynamics for Reliable Hallucination Detection in LLMs}

\author{
 \textbf{Zhenliang Zhang\textsuperscript{1,2}},\ 
 \textbf{Xinyu Hu\textsuperscript{1}},\ 
 \textbf{Huixuan Zhang\textsuperscript{1}}
 \\
 \textbf{Junzhe Zhang\textsuperscript{1}},\ 
 \textbf{Xiaojun Wan\textsuperscript{1}}
\\
 \textsuperscript{1}Wangxuan Institute of Computer Technology, Peking University
 \\
 \textsuperscript{2}School of Software and Microelectronics, Peking University
\\
 {\texttt{\{zhenliang,zhanghuixuan,junzhezhang\}@stu.pku.edu.cn}}
 \\
 {\texttt{\{huxinyu,wanxiaojun\}@pku.edu.cn}}
}

\begin{document}
\maketitle
\begin{abstract}

Large language models (LLMs) excel at various natural language processing tasks, but their tendency to generate hallucinations undermines their reliability. Existing hallucination detection methods leveraging hidden states predominantly focus on static and isolated representations, overlooking their dynamic evolution across layers, which limits efficacy. To address this limitation, we shift the focus to the hidden state update process and introduce a novel metric, the ICR Score (\textbf{I}nformation \textbf{C}ontribution to \textbf{R}esidual Stream), which quantifies the contribution of modules to the hidden states' update. We empirically validate that the ICR Score is effective and reliable in distinguishing hallucinations. Building on these insights, we propose a hallucination detection method, the ICR Probe, which captures the cross-layer evolution of hidden states. Experimental results show that the ICR Probe achieves superior performance with significantly fewer parameters. Furthermore, ablation studies and case analyses offer deeper insights into the underlying mechanism of this method, improving its interpretability.

\end{abstract}

\section{Introduction}

Large language models (LLMs) demonstrate remarkable performance across various natural language processing tasks \cite{zhang2023siren}. However, these models are still prone to generating hallucinations, which are nonsensical or irrelevant content that deviates from the intended output \cite{10.1145/3571730}. This issue highlights the critical need for effective methods of hallucination detection.

Various methods for hallucination detection exist. Mainstream approaches analyze generated output through consistency checks or reference comparisons \cite{manakul-etal-2023-selfcheckgpt}, while probability-based methods focus on logit probability uncertainty \cite{renOutofDistributionDetectionSelective2022}. Another approach examines hidden states (e.g., embedding vectors) across LLM layers to detect hallucinations \cite{chen2024inside, 79054f3cfa7d4fb0a9e172da20ba04a2}. Methods based on output or logit probabilities often require ground truth references or multiple generations for consistency. In contrast, hidden state based detection offers the advantage of being reference-free, eliminating the need for external sources.

Current hallucination detection methods based on hidden states can be broadly categorized into training-based and training-free methods. Training-based methods often involve training individual or combination probes \cite{azariaInternalStateLLM2023, ch-wang-etal-2024-androids}, or using semantic entropy probes \cite{Kossen2024SemanticEP}, whereas training-free methods calculate detection metrics directly from hidden states \cite{sriramananLLMCheckInvestigatingDetection2024a}. 
However, existing methods typically focus on static, high-dimensional hidden states (around 4000 dimensions), which limits feature extraction capabilities. These methods make it challenging to capture the updates of hidden states and the cross-layer evolution of the residual stream, ultimately restricting the effectiveness of hallucination detection.

To overcome these limitations, we introduce a novel approach that \textbf{shifts the focus from the hidden states themselves to their update process across layers}. Specifically, the \textbf{ICR} Score (\textbf{I}nformation \textbf{C}ontribution to \textbf{R}esidual Stream) quantifies the contribution of different modules (e.g., FFN or self-attention) to hidden state updates at each layer. Empirical results demonstrate that the ICR Score captures a stable and consistent pattern of residual stream updates, showing strong potential for distinguishing hallucinations.

We further introduce the \textbf{ICR Probe}, which aggregates ICR Scores across all layers to capture the comprehensive dynamics of the residual stream. 
Experimental results on three mainstream open-source LLMs demonstrate that the ICR Probe effectively detects hallucinations and outperforms previous methods across multiple datasets. Additionally, ablation studies are conducted to reveal the underlying mechanisms.
In summary, our core contributions are:
\begin{itemize} 
\item \textbf{Novel Detection Signal}: We propose a hallucination detection method by focusing on the update patterns of hidden states across layers, introducing the ICR Score, a metric that captures the dynamic evolution of residual stream updates.

\item \textbf{ICR Probe Development}: We introduce the ICR Probe, an effective and robust tool for hallucination detection, offering superior performance with fewer parameters. 

\item\textbf{Empirical Validation}: We conduct extensive empirical evaluations, showcasing the effectiveness, generalizability, and interpretability of our method across various datasets, reinforcing the understanding of its underlying mechanisms\footnote{The code is available in \url{https://github.com/XavierZhang2002/ICR_Probe}}.
\end{itemize}

\section{Background}
\label{sec:background}
 
\paragraph{Information Contribution of Different Modules} In each layer of large language models (LLMs), multi-head self-attention (MHSA) and feed-forward networks (FFN) perform distinct but complementary functions in information processing and knowledge integration \cite{geva-etal-2021-transformer}.

\noindent MHSA functions as a \textbf{contextual signal router}, dynamically modulating token interactions based on query-key affinity scores. By attending to preceding tokens, MHSA redistributes existing contextual information and integrates it into the evolving residual stream of hidden states \cite{elhage2021mathematical}. Importantly, MHSA does not directly extract new knowledge from the model parameters. Instead, it reallocates information already present in token representations, enhancing relevant associations within the context.

\noindent In contrast, FFNs act as \textbf{key-value memory banks}, facilitating sparse, content-based knowledge retrieval \cite{geva-etal-2023-dissecting}. 
Unlike MHSA, FFNs do not extract new information from the context but rather retrieve and integrate knowledge stored in the model's internal parameters, injecting learned knowledge into the residual stream.

\paragraph{Layer-wise Information Contribution Characteristics}

In LLMs, the contributions of FFNs and MHSA vary across layers, leading to distinct layer-specific characteristics. \citet{stolfo-etal-2023-mechanistic} employ causal mediation analysis to investigate the information flow in LLMs during reasoning tasks. Their analysis reveals that FFNs predominantly influence the early and deeper layers, while MHSA plays a central role in the intermediate layers. Specifically, FFNs drive the processing of operands and operators, while MHSA manages the dynamic redistribution of information across tokens. In later stages, FFNs retrieve and integrate task-specific knowledge to generate results.
Similar findings are reported by \citet{elhage2021mathematical} and \citet{geva-etal-2023-dissecting}, who observe that FFNs and MHSA dominate hidden state updates at different layers. These differences in information contribution provide valuable insights into the model’s internal information flow patterns.

\begin{figure*}[!t]
\centering
\includegraphics[width=0.95\textwidth]{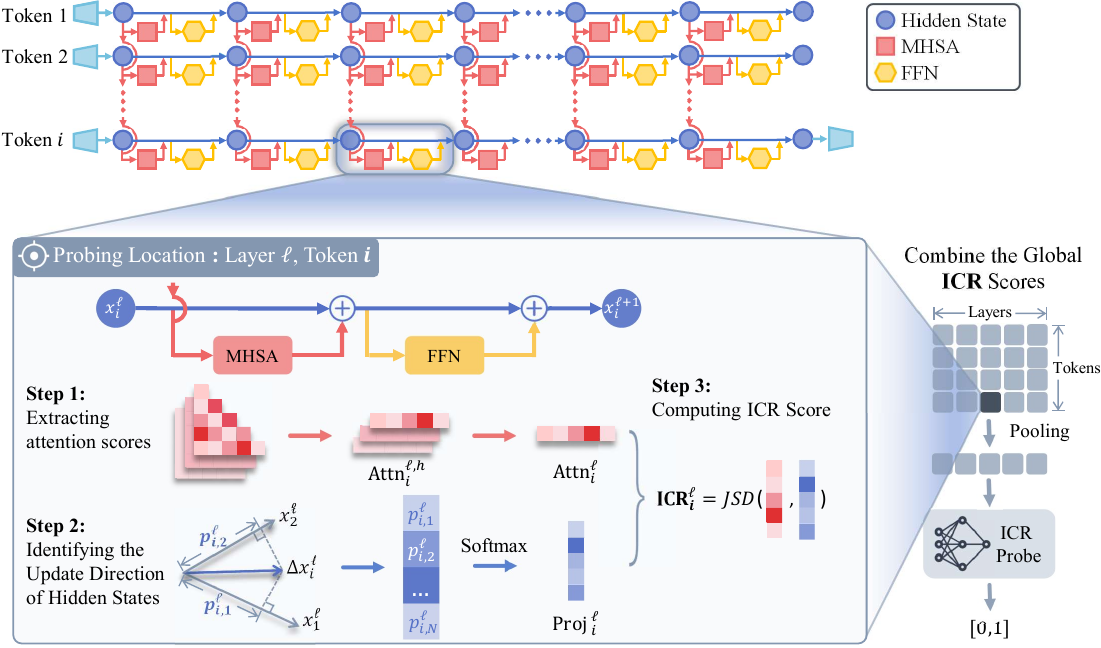}
\caption{Overview of the ICR Score computation and ICR Probe detection process. We first probe each layer's residual stream to obtain the ICR Score, then aggregate the ICR Scores across all layers to capture global information. This aggregated signal serves as the input to the probe, which produces the final hallucination detection result.}
\label{fig:main_figure}
\end{figure*}

\section{ICR Score}

Previous research by \citet{stolfo-etal-2023-mechanistic} suggests that the information contribution to the residual stream update process may be correlated with model performance, making it a potential signal for hallucination detection. Based on this insight, we formalize the residual stream and introduce the \textbf{ICR} Score (\textbf{I}nformation \textbf{C}ontribution to \textbf{R}esidual Stream), a metric that quantifies each module’s contribution to the residual stream updates.

\subsection{Residual Stream Update Process}

We describe the residual stream update process, focusing on how the varying contributions of different modules lead to distinct characteristics across layers.

\paragraph{Residual stream update process in LLMs.}

LLMs process inputs through multiple sequential computational stages. First, the input text is tokenized into \( N \) discrete tokens \( \{t_i\}_{i=1}^N \). Each token is then mapped to a \( d \)-dimensional hidden state.
These hidden states evolve iteratively through the \( L \) layers of decoder architecture. At layer \( \ell \), the hidden states are updated as shown in Equation~\eqref{eq:info_stream}, where \( x^\ell_i \) is the hidden state of token \( t_i \) in layer \( \ell \):
\begin{equation}
\small
    x^\ell_i = x^{\ell-1}_i + \underbrace{\text{MHSA}^\ell(x^{\ell-1}_i)}_{a^\ell_i} + \underbrace{\text{FFN}^\ell(x^{\ell-1}_i + a^\ell_i)}_{m^\ell_i}
    \label{eq:info_stream}
\end{equation}
The hidden state is updated through two key modules: multi-head self-attention (MHSA) and feedforward networks (FFN). 

\noindent \(a^\ell_i\) represents the contribution from the MHSA module, and \(m^\ell_i\) represents the contribution from the FFN module. As discussed in Section \ref{sec:background}, contemporary research highlights the differing roles of these two modules in processing information.

\paragraph{Layer-wise Characteristics of Residual Stream Updates.} %

As discussed in Section~\ref{sec:background}, the updates to the residual stream of hidden states differ across layers, with each layer contributing uniquely through the MHSA and FFN modules. 

\subsection{Constructing the ICR Score}

To quantify the contribution of different modules to the updates of hidden states, we define the total update of hidden states at layer \( \ell \) as:

\begin{equation}
\Delta x^\ell_i = x^\ell_i - x^{\ell-1}_i = a^\ell_i + m^\ell_i
\label{eq:delta_x}
\end{equation}

Here, \( \Delta x^\ell_i \) represents the update of the hidden state, which consists of two components: the contextual information \( a^\ell_i \) derived from the MHSA module, and the parametric knowledge \( m^\ell_i \) derived from the FFN module. Our objective is to quantify the specific contribution of each module.

A straightforward method for quantifying this contribution is to compute the ratio of the respective contributions in the residual stream, such as \( \frac{a^\ell_i}{a^\ell_i + m^\ell_i} \). However, this ratio may not always be reliable, since \( m^\ell_i \) is computed using \( a^\ell_i \), which complicates the distinction between the FFN's contribution of new parameterized knowledge and its reinforcement of information from the MHSA.

To address this issue, we propose an effective method for quantifying the contributions by measuring the consistency between the hidden state update \( \Delta x^\ell_i \) and the attention score. As illustrated in Figure~\ref{fig:main_figure}, the computation process consists of three steps:

\paragraph{Extracting Attention Scores}

For the \( i \)-th token in layer \( \ell \), the attention score with respect to the \( j \)-th token in the context for a given attention head \( h \) is computed as:

\begin{equation}
\small
\text{Attn}_{i,j}^{\ell,h} = \frac{(Q_i^h)^T \cdot K_j^h}{\sqrt{d}}
\label{eq:attn_score}
\end{equation}
where \( Q_i^h \) and \( K_j^h \) are the query and key vectors for the \( i \)-th and \( j \)-th tokens, and \( d \) is the dimension of the vectors.

\noindent The attention score vector for the \( i \)-th token across all \( N \) tokens is:

\begin{equation}
\small
\text{Attn}_i^{\ell,h} = \left[\text{Attn}_{i,1}^{\ell,h}, \text{Attn}_{i,2}^{\ell,h}, \dots, \text{Attn}_{i,N}^{\ell,h}\right] \in \mathbb{R}^N
\label{eq:attn_score_list}
\end{equation}

\noindent To obtain a more robust attention score for the \( i \)-th token, we average the scores across all \( H \) attention heads:

\begin{equation}
\small
\text{Attn}_i^{\ell} = \frac{1}{H} \sum_{h=1}^{H} \text{Attn}_i^{\ell,h} \in \mathbb{R}^N
\label{eq:attn_score_pooling}
\end{equation}

\paragraph{Identifying the Update Direction of Hidden States}

The set \( \{x^\ell_j\}_{j=1}^{N} \) represents the hidden states of all \( N \) input tokens. To quantify how much the update \( \Delta x^\ell_i \)  incorporates information from the context, we calculate the projection of \( \Delta x^\ell_i \) onto the hidden states \( x^\ell_j \) of all tokens \( \{t_i\}_{i=1}^N \) in the input context.   Specifically, we calculate the projection length \( p_{i,j}^\ell \) between \( \Delta x^\ell_i \) and \( x^\ell_j \), normalized by the norm of \( x^\ell_j \). This is given by Equation~\ref{eq:projection}.

\begin{equation}
\small
p_{i,j}^\ell = \frac{(\Delta x^\ell_i)^T \cdot x^\ell_j}{\| x^\ell_j \|}
\label{eq:projection}
\end{equation}

\noindent where \( \| x^\ell_j \| \) is the magnitude (norm) of the hidden state of the \( j \)-th token. The projection length \( p_{i,j}^\ell \) measures the magnitude of the update \( \Delta x^\ell_i \) in the direction of the hidden state \( x^\ell_j \). 

We then construct the projection vector \(\text{Proj}_i^\ell = [p_{i,j}^\ell]_{j=1}^N\) and normalize it using softmax.

\paragraph{Computing Consistency Between Hidden State Update Direction and Attention Scores}

We measure the consistency between the hidden state update direction \( \text{Proj}_i^\ell \) and the attention scores \( \text{Attn}_i^\ell \) using the Jensen-Shannon Divergence (JSD). The \textbf{ICR} Score (\textbf{I}nformation \textbf{C}ontribution to \textbf{R}esidual Stream) is defined in Equation~\ref{eq:icore_score}, with the top \( k \) tokens selected based on the highest attention scores. Appendix~\ref{appendix:icr_score} provides more details.
\begin{equation}
\text{ICR}_i^\ell = \text{JSD}( \text{Proj}_i^\ell, \text{Attn}_i^\ell )
\label{eq:icore_score}
\end{equation}

\noindent The ICR score quantifies each module's contribution to the residual stream update at each layer, identifying the dominant module shaping the hidden state update \( \Delta x^\ell_i \).  The interpretations of the ICR score are as follows:
\begin{itemize}
    \item A small ICR score indicates that the hidden state update \( \Delta x^\ell_i \) aligns closely with the attention scores \( \text{Attn}_i^\ell \). This implies that the \textbf{attention mechanism predominantly drives the hidden state update}, while the FFN primarily reinforces the information without adding significant new content.
    \item A large ICR score signifies a noticeable divergence between the hidden state update \( \Delta x^\ell_i \) and the attention scores. \textbf{This divergence suggests that the FFN is more dominant in determining the update}, with the attention mechanism contributing less to the information stream.
\end{itemize}
The ICR score captures the relative contribution of each module across \( L \) layers and \( N \) tokens, revealing how attention and FFN modules shape the model’s residual stream.

\section{ICR Probe for Hallucination Detection}

In this section, we analyze the layer-wise characteristics of the ICR Score and its ability to distinguish hallucinations, and design the ICR Probe for hallucination detection.

\begin{figure}[!t]
\centering
\includegraphics[width=0.9\columnwidth]{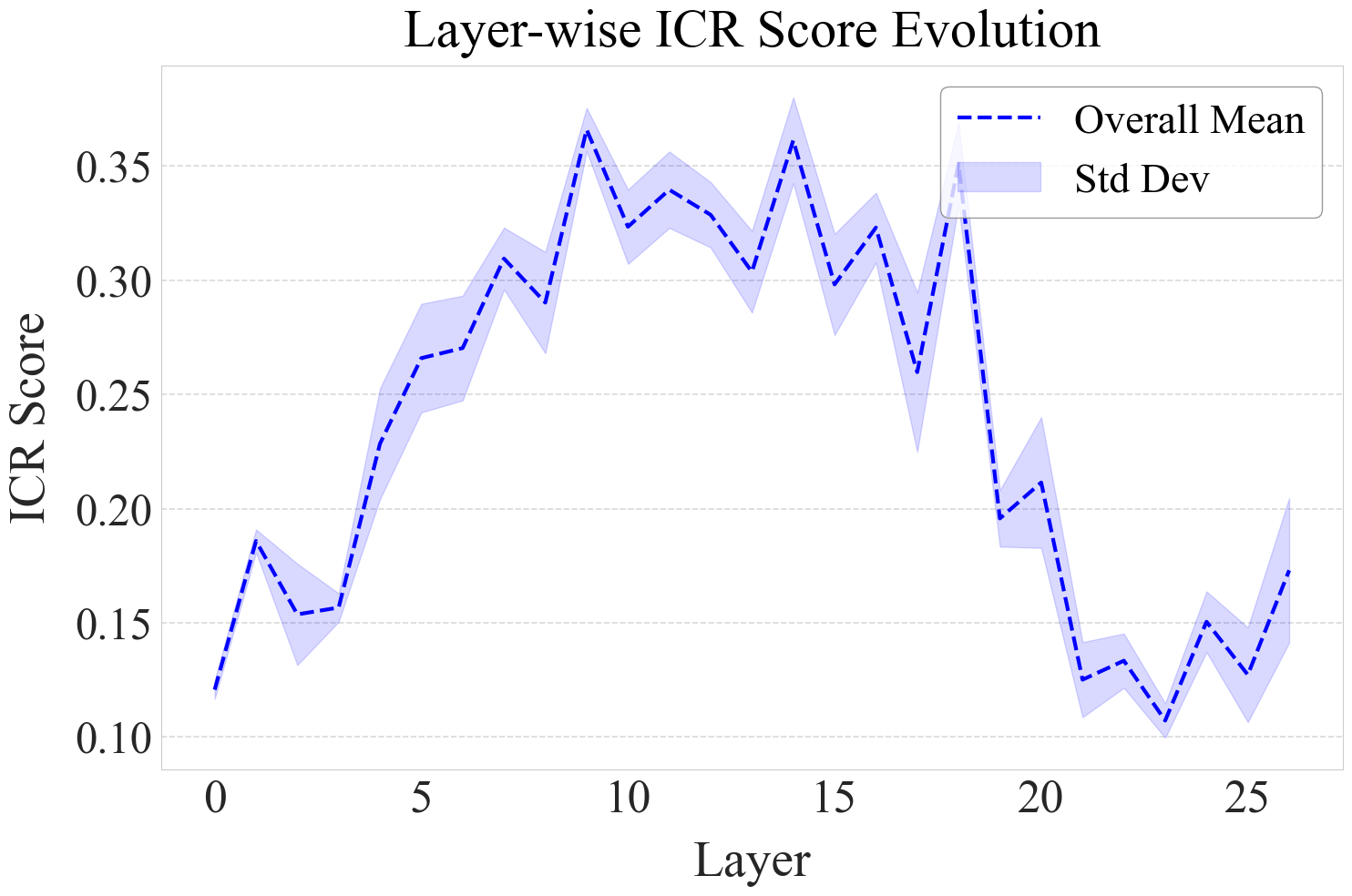}
\caption{Mean ICR scores with standard deviation bands across four datasets. The narrow standard deviation bands indicate that the ICR score captures the residual stream patterns with strong cross-dataset consistency and stability.}
\label{fig:icore_layerwise_analysis}
\end{figure}

\begin{figure*}[!t]
\centering 
\begin{subfigure}[b]{0.46\textwidth}
    \centering
    \includegraphics[width=\textwidth]{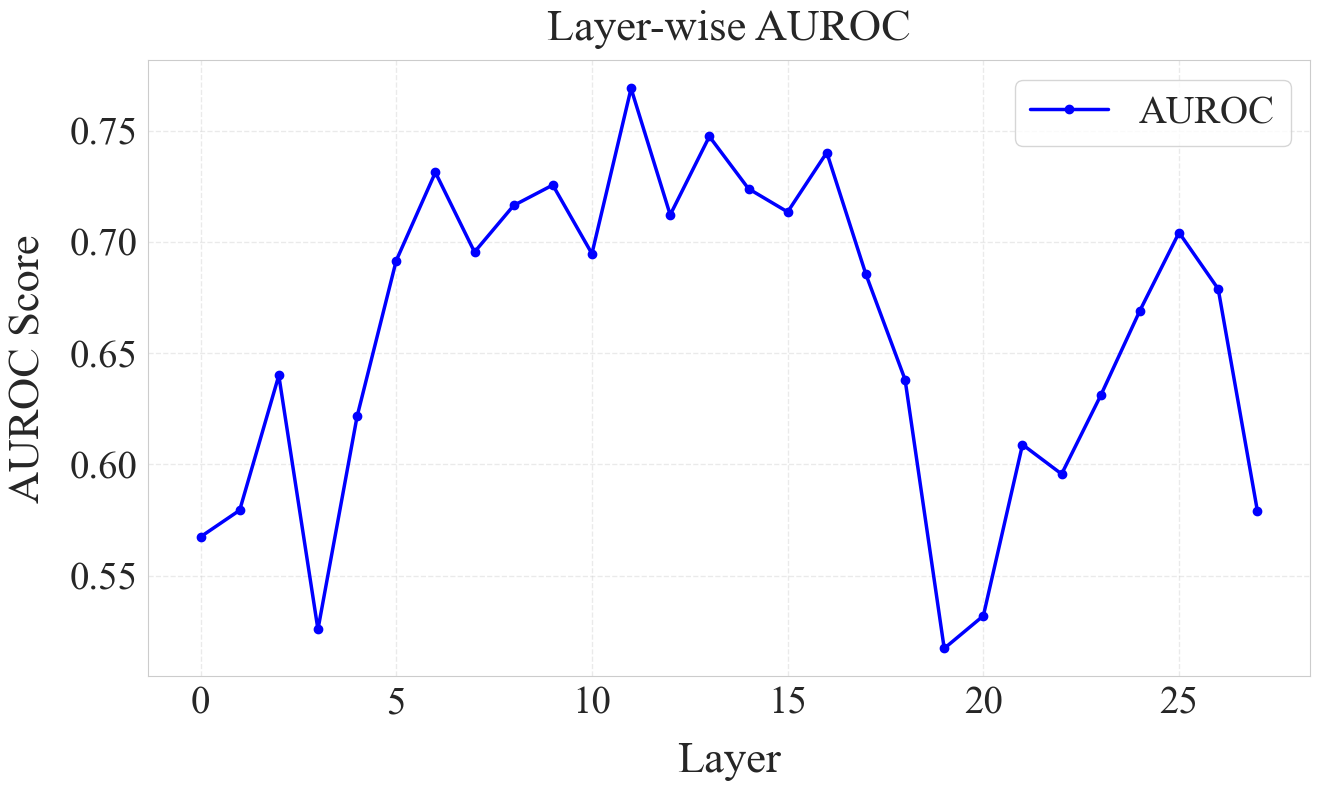}
    \caption{Layer-wise AUROC for hallucination detection \textbf{using direct ICR scores}, peaking at layer \textbf{11}, which indicates the layer most effective for hallucination detection.}
    \label{fig:layerwise_auroc}
\end{subfigure}
\hspace{5mm} 
\begin{subfigure}[b]{0.46\textwidth}
    \centering
    \includegraphics[width=\textwidth]{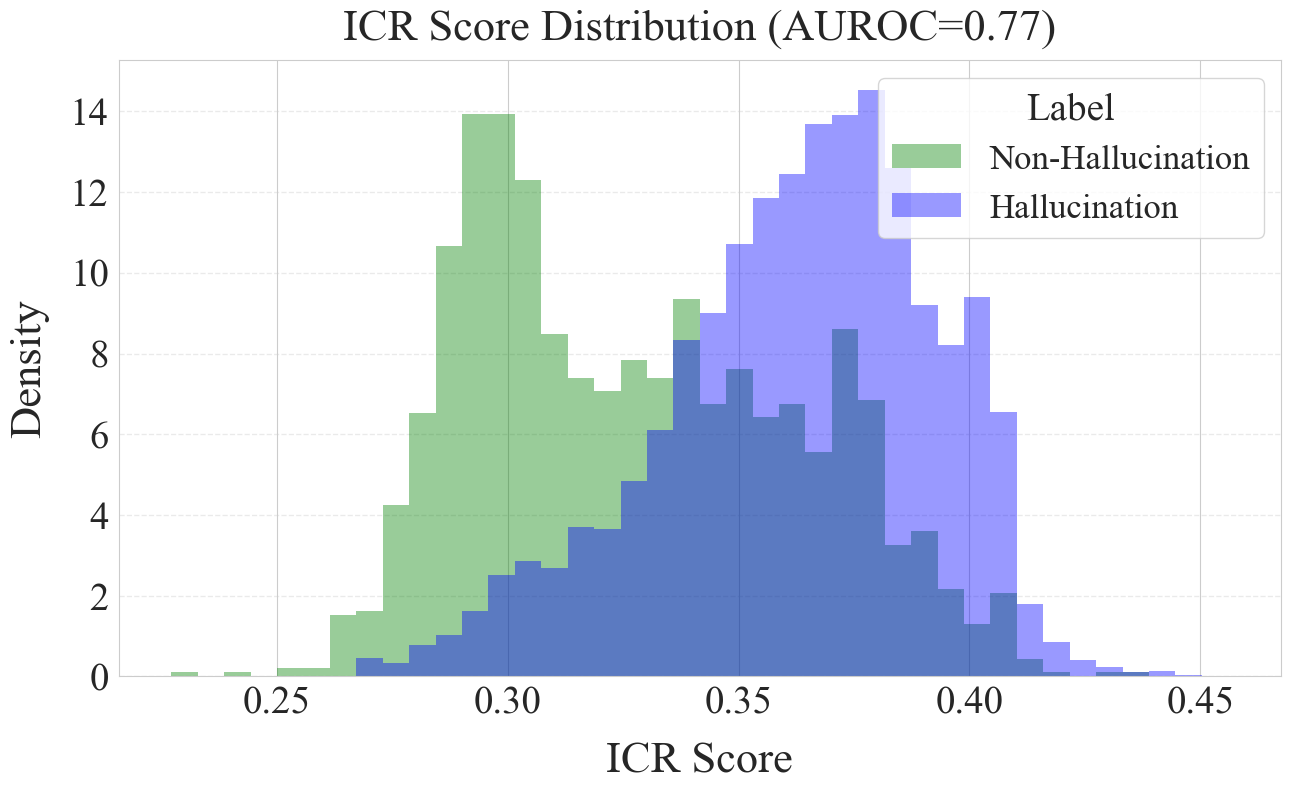}
    \caption{Probability density distribution of ICR Scores at layer 11, showing clear separability between hallucinated and non-hallucinated samples. }
    \label{fig:icore_density}
\end{subfigure}

\caption{ICR Score for hallucination detection: (a) Layer-wise AUROC and (b) probability density distributions, highlighting ICR's effectiveness as a predictive metric.}
\label{fig:icore_difference}
\end{figure*}

\subsection{Empirical Study of the ICR Score}
\label{sec:empirical_study}
We investigate the potential of the ICR score for hallucination detection, addressing two requirements: (1) consistency/stability of layer-wise features, and (2) its ability to distinguish between hallucinated and non-hallucinated outputs.

\paragraph{Stability of Layer-wise Features in the ICR Score}

Using Qwen2.5 as an example, we compute the mean and variance of the ICR scores across layers on four different datasets (model and dataset details in Section~\ref{sec:exp_setup}) to examine the layer-wise characteristics and cross-dataset consistency.

\noindent As shown in Figure~\ref{fig:icore_layerwise_analysis}, the ICR score follows a clear progression across layers. In the early layers (0–3), low scores indicate that updates are primarily driven by MHSA, focusing on local contextual extraction. From layers 4–20, the ICR score increases, peaking around layers 10–15, suggesting a shift towards FFN-dominated updates, where parametric knowledge is integrated. Beyond layer 21, the score declines, signaling a return to MHSA-driven updates for refining contextual integration.

\noindent The narrow standard deviation bands in the figure further support the consistency of this pattern. This feature remains \textbf{consistent across datasets}, suggesting that the ICR score reflects an intrinsic property of the model rather than dataset-specific variations. It provides a structured representation of the information flow within LLMs.

\paragraph{\textbf{Takeaway}} The ICR score systematically maps the model's intrinsic residual stream update pattern, providing a stable feature that confirms its potential as a diagnostic tool for understanding model performance.


\paragraph{ICR Score’s Ability to Distinguish Hallucination}

To evaluate the detection capability of the ICR score, we assess its ability to distinguish hallucinated outputs from non-hallucinated ones. The ICR score is used as a direct feature for classification. Using the HaluEval dataset as an example, Figure~\ref{fig:layerwise_auroc} presents the AUROC values across layers, illustrating the inherent separability of ICR for hallucination detection. AUROC exceeds 0.7 in ten layers, peaking at 0.7690 at layer 11, which indicates that the ICR score effectively reflects the model’s performance trends.

\noindent Figure~\ref{fig:icore_density} shows the probability density distribution of ICR scores at layer 11, with clear separation between hallucinated and non-hallucinated samples, confirming ICR's ability to distinguish reliable from erroneous generations.

\paragraph{\textbf{Takeaway}} The ICR score effectively distinguishes between hallucinated and non-hallucinated outputs, confirming its potential as a predictive metric for hallucination detection.

\begin{table*}[t]
\centering
\small
\setlength{\tabcolsep}{10pt} 
\begin{tabular}{@{}c|c|cccc@{}}
\toprule
\textbf{LLMs} & \textbf{Methods} & \textbf{HaluEval} & \textbf{SQuAD} & \textbf{HotpotQA} & \textbf{TriviaQA} \\ \midrule
\multirow{6}{*}{\textbf{Gemma-2}} & PPL & 0.5546 & 0.5419 & 0.7196 & 0.7765 \\
 & LN-Entropy & 0.7451 & 0.6531 & 0.7172 & 0.7200 \\
 & LLM-check & 0.5683 & 0.5713 & 0.5600 & 0.5821 \\
 & SAPLMA & \underline{0.8101} & \underline{0.7175} & \underline{0.8193} & 0.7751 \\
 & SEP & 0.6429 & 0.6417 & 0.6356 & \underline{0.7834} \\
 & \textbf{ICR Probe (Ours)} & \textbf{0.8436} & \textbf{0.8142} & \textbf{0.8409} & \textbf{0.8001} \\ \midrule
\multirow{6}{*}{\textbf{Qwen2.5}} & PPL & 0.5439 & 0.5303 & 0.6125 & 0.6974 \\
 & LN-Entropy & 0.7371 & 0.6497 & 0.6855 & 0.7004 \\
 & LLM-check & 0.5292 & 0.5700 & 0.5445 & 0.5552 \\
 & SAPLMA & \underline{0.7799} & \underline{0.6929} & \underline{0.7750} & \textbf{0.8225} \\
 & SEP & 0.6578 & 0.6363 & 0.6470 & 0.7520 \\
 & \textbf{ICR Probe (Ours)} & \textbf{0.8003} & \textbf{0.7456} & \textbf{0.7917} & \underline{0.7684} \\ \midrule
\multirow{6}{*}{\textbf{Llama-3}} & PPL & 0.5920 & 0.6399 & 0.6721 & 0.7012 \\
 & LN-Entropy & 0.6508 & 0.6306 & 0.6593 & 0.5860 \\
 & LLM-check & 0.5206 & 0.5411 & 0.5491 & 0.5417 \\
 & SAPLMA & 0.7238 & 0.7107 & \underline{0.7701} & \textbf{0.7650} \\
 & SEP & \underline{0.7378} & \underline{0.7201} & 0.6541 & 0.7116 \\
 & \textbf{ICR Probe (Ours)} & \textbf{0.7603} & \textbf{0.7634} & \textbf{0.7982} & \underline{0.7325} \\ \bottomrule
\end{tabular}
\caption{Hallucination detection performance on four datasets. Higher AUROC values indicate better performance.}
\label{tab:main_results}
\end{table*}

\subsection{ICR Probe}
\label{sec:icr_probe}

The previous analysis demonstrates that the ICR Score effectively captures the model’s information contributions and serves as a reliable indicator for hallucination detection. However, utilizing the ICR score from a single layer provides only a partial representation of residual stream updates. To capture the global update pattern of the residual stream, we introduce the \textbf{ICR Probe}, a classifier trained on ICR Scores across all layers to detect hallucinations.

The ICR Probe takes the averaged ICR score as input, which is obtained by performing token-wise averaging of the original \( N \times L \) matrix, resulting in a \( 1 \times L \) vector. The output is a scalar value between 0 and 1, representing the likelihood of a non-hallucinated generation.

\paragraph{Architecture}  
Based on empirical evaluations, we adopt the model architecture \( (L, 128, 64, 32, 1) \) for the ICR Probe, which optimally balances model complexity with computational efficiency. For models with \( L < 42 \), this configuration ensures that the total number of parameters remains below \textbf{16K}, maintaining computational efficiency. Detailed specifications are provided in Appendix~\ref{appendix:probe}.

\section{Experiments}

We assess the performance of the ICR Probe for hallucination detection, comparing it with baselines. Additionally, we investigate its generalizability across different datasets, conduct ablation studies, and present a case study focused on token-level detection.

\subsection{Settings}
\label{sec:exp_setup}

\paragraph{Models}
We evaluate the detection performance of ICR Probe on three mainstream open-source LLMs: Llama-3-8B-Instruct \cite{llama3modelcard}, Qwen2.5-7B-Instruct \cite{qwen2.5}, and Gemma-2-9B-it \cite{gemmateam2024gemma2improvingopen}.

\paragraph{Datasets}
The models are evaluated on several benchmark datasets: HaluEval \cite{li-etal-2023-halueval}, which focuses on hallucination detection; SQuAD \cite{rajpurkar-etal-2016-squad}, a reading comprehension dataset; TriviaQA \cite{joshi-etal-2017-triviaqa}, which tests general knowledge; and HotpotQA \cite{yang-etal-2018-hotpotqa}, designed for multi-hop question answering.

\paragraph{Baselines}

We compare hallucination detection performance with several baselines. The training-free methods include \textbf{PPL} \cite{renOutofDistributionDetectionSelective2022}, \textbf{LN-Entropy} \cite{malininUncertaintyEstimationAutoregressive2020}, and \textbf{LLM-check} \cite{sriramananLLMCheckInvestigatingDetection2024a}. The training-based methods include \textbf{SAPLMA} \cite{azariaInternalStateLLM2023} and \textbf{SEP} \cite{Kossen2024SemanticEP}. Details of these baselines are in Appendix~\ref{appendix:baselines}.

\paragraph{Evaluation Metric}
Since hallucination detection is a binary classification task, we use the Area Under the Receiver Operating Characteristic Curve (AUROC), a threshold-independent metric that captures the trade-off between true and false positive rates.

\paragraph{ICR Probe Training}
The ICR Probe is trained on the pooled ICR Score to detect hallucinations. It is a lightweight multi-layer perceptron (MLP) classifier with four fully connected layers. The training follows a standard supervised learning setup with binary cross-entropy loss and Adam optimization. 

Following the experimental setup outlined by \citet{orgad2024llmsknowshowintrinsic}, we partition each dataset into training and testing sets and report the results for each dataset accordingly. \textbf{The experimental setup for baselines matches ours exactly.} Additionally, we conduct a dedicated evaluation of generalization in Section~\ref{generalization}.

\begin{figure*}[!t]
\centering
\includegraphics[width=1\textwidth]{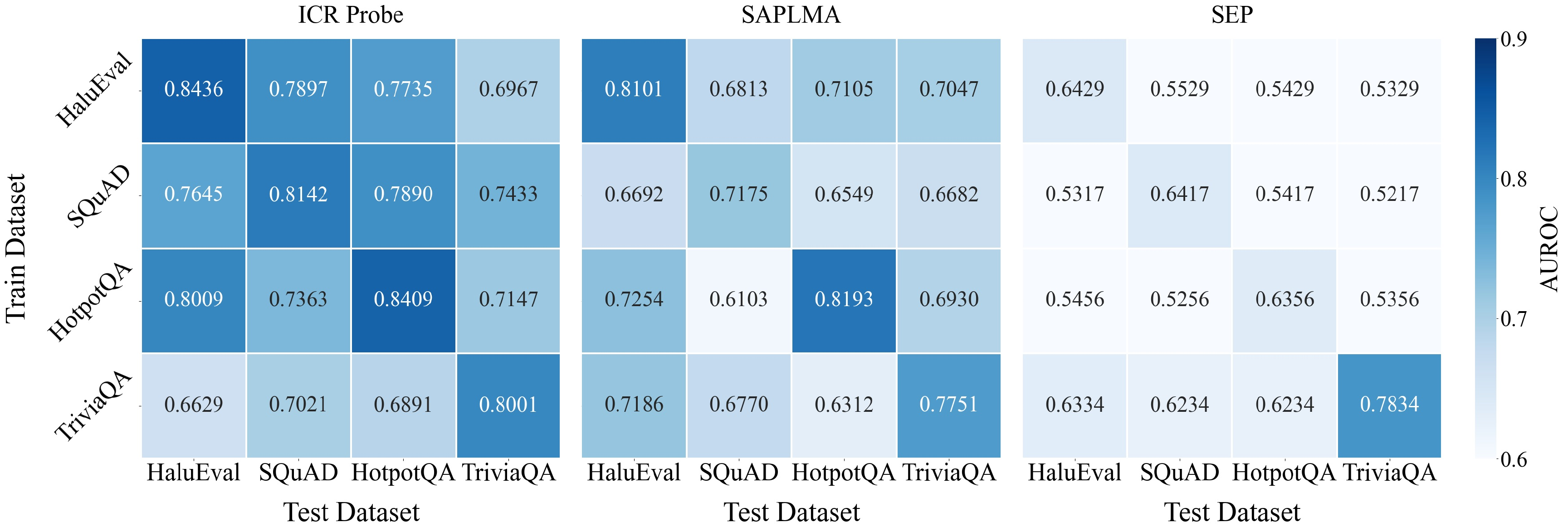}
\caption{Cross-dataset generalization heatmaps for ICR Probe, SAPLMA, and SEP. Each subplot displays the AUROC when the model is trained on the row dataset and tested on the column dataset, with values annotated in each cell. A shared color scale highlights comparative generalization performance, showing that the ICR Probe not only achieves higher absolute AUROCs across unseen datasets but also exhibits the smallest average drop under domain shift.}
\label{fig:generalization}
\end{figure*}

\subsection{Hallucination Detection Results}

The performance of the ICR Probe is evaluated across four datasets and three mainstream open-source LLMs. Table~\ref{tab:main_results} presents the key experimental results, showing that the ICR Probe outperforms both train-free and training-based baseline methods on most datasets.

Our approach offers several distinct advantages beyond its effectiveness in hallucination detection. First, the ICR Probe integrates global information from all layers, unlike methods that focus on isolated hidden states, which capture only local information. Our method also incorporates the contribution of modules, capturing the entire global residual stream update process. This provides a more comprehensive view of the model's behavior. Additionally, the ICR Probe eliminates the need to select specific layers or tokens, making it highly adaptable and scalable across different models and tasks. It also enables real-time hallucination detection from a single generation, eliminating the need for multiple samples. Furthermore, the ICR Probe is parameter efficient, with only 16K parameters, significantly smaller than alternatives like SAPLMA, which has 110K parameters \cite{azariaInternalStateLLM2023}. This ensures the approach remains both scalable and practical.

\subsection{Generalization Analysis}
\label{generalization}

We assess cross-dataset generalization by training each model on one dataset and evaluating on another, visualized as heatmaps in Figure~\ref{fig:generalization}. Each cell reports the AUROC of the model when trained on the row dataset and tested on the column dataset, using a shared color scale (0.6–0.9) for direct comparison.

Across unseen datasets, our ICR Probe maintains AUROC $\geq$ 0.66 and outperforms both SAPLMA and SEP. Furthermore, the average AUROC drop from in-domain to cross-domain is only 8.61 \% for our method, versus 10.18 \% for SAPLMA and 11.67 \% for SEP, underscoring its superior accuracy and robustness under domain shift.

This advantage stems from leveraging layer-wise residual stream evolution patterns-capturing intrinsic LLM dynamics independent of dataset-specific artifacts (see Figures~\ref{fig:icore_layerwise_analysis} and~\ref{fig:icr_score_appendix})---whereas SAPLMA and SEP rely on data---dependent representations and thus exhibit greater cross-domain sensitivity.


\begin{table}[t]
\centering
\small
\setlength{\tabcolsep}{2pt} 
\begin{tabular}{@{}cl|cccc@{}}
\toprule
\multicolumn{2}{c|}{\textbf{Setting}} & \multirow{2}{*}{\textbf{HaluEval}} & \multirow{2}{*}{\textbf{SQuAD}} & \multirow{2}{*}{\textbf{HotpotQA}} & \multirow{2}{*}{\textbf{TriviaQA}} \\
\multicolumn{1}{l}{\textbf{$\text{Attn}_i^{\ell}$}} & \textbf{$\text{Proj}_i^{\ell}$} &  &  &  &  \\ \midrule
$\times$ & $\times$  & 0.5000 & 0.5000 & 0.5000 & 0.5000 \\
$\times$ & \checkmark & 0.7993 & 0.7765 & 0.7829 & 0.7723 \\
\checkmark & \checkmark & 0.8436 & 0.8142 & 0.8409 & 0.8001 \\ \bottomrule
\end{tabular}
\caption{Ablation study of the ICR score components, showing the impact of attention (\(\text{Attn}_i^\ell\)) and hidden state projections (\(\text{Proj}_i^\ell\)) on hallucination detection performance.}
\label{tab:ablation_results}
\end{table}

\subsection{Ablation Experiments and Interpretability}

We conduct ablation experiments on Gemma-2 to investigate two main aspects: (1) the contribution of different components of the ICR Score to hallucination detection, and (2) the impact of different layers on the Probe's performance.

\paragraph{Ablation Study 1: Contribution of Each component.}

As defined in Equation~\ref{eq:icore_score}, the ICR Score consists of two components: the update directions of hidden states (\(\text{Proj}_i^\ell\)) and the attention scores (\(\text{Attn}_i^\ell\)). To evaluate the contribution of each component, we design three experimental settings:
In the NONE setting, both the hidden state (\(\text{Proj}_i^\ell\)) and attention (\(\text{Attn}_i^\ell\)) signals are excluded, resulting in random performance (AUROC = 0.5). In the HS ONLY setting, only the hidden state information (\(\text{Proj}_i^\ell\)) is used, while the attention scores (\(\text{Attn}_i^\ell\)) are set to a uniform distribution. At this stage, the calculation corresponds to the entropy of the \(\text{Proj}_i^\ell\). Finally, in the HS + ATTN setting, both the hidden state and attention signals (\(\text{Proj}_i^\ell\) and \(\text{Attn}_i^\ell\)) are integrated. Details of the ablation experiments are provided in Appendix~\ref{appendix:ablation}.

\noindent The results, shown in Table~\ref{tab:ablation_results}, demonstrate that even the hidden state component alone significantly improves hallucination detection. This suggests that the information entropy of \(\text{Proj}_i^\ell\) is critical. Moreover, combining both components (hidden state and attention) improves performance, highlighting their complementary nature and validating the construction of the ICR Score.

\paragraph{Ablation Study 2: Layer Contribution Analysis.}

\begin{table}[t]
\centering
\small
\setlength{\tabcolsep}{2.3pt} 
\begin{tabular}{@{}l|cccc@{}}
\toprule
\textbf{Setting} & \textbf{HaluEval} & \textbf{SQuAD} & \textbf{HotpotQA} & \textbf{TriviaQA} \\ \midrule
\textbf{Full Layers} & 0.8436 & 0.8142 & 0.8409 & 0.8001  \\ \midrule
\multicolumn{1}{l|}{w/o \text{EL}} & 0.7833 & 0.8045 & 0.8132 & 0.7885 \\
\multicolumn{1}{l|}{w/o \text{ML}} & 0.7774 & 0.7582 & 0.7421 & 0.7197 \\
\multicolumn{1}{l|}{w/o \text{DL}} & 0.7285 & 0.7924 & 0.7936 & 0.7657 \\ \bottomrule
\end{tabular}
\caption{Layer-wise ablation study showing AUROC scores under different layer configurations.}
\label{tab:layer_ablation_results}
\end{table}

We perform a layer-wise ablation study to analyze the contributions of different layer groups in the LLM to hallucination detection. The model is divided into three stages: Early Layers (Layer 0-13), Middle Layers (Layer 15-28), and Deep Layers (Layer 29-41). In each setting, \textbf{remove} a specific layer group, e.g. \text{Early Layers} (EL), \text{Middle Layers} (ML), \text{Deep Layers} (DL). 

\noindent The results in Table~\ref{tab:layer_ablation_results} show that removing Middle Layers significantly reduces performance, emphasizing their critical role in hallucination detection.

\subsection{Token-level Detection: Case Study}

Although our method is designed for sequence-level detection (as discussed in Section~\ref{sec:icr_probe}), we perform token-level analysis to explain the probe's detection capabilities and characteristics.

In token-level detection, each token is processed individually by the ICR Probe, with the prediction probability indicating its correctness. This method is effective only for detecting hallucinations in \textbf{key tokens} that are crucial to the answer. Table~\ref{tab:token_level_detection} shows that in hallucinated cases, key tokens like \texttt{hectares} and \texttt{Vietnam} have low probabilities (\(0.04\) and \(0.19\)), successfully identifying hallucinations, while common tokens like \texttt{The} and \texttt{.} have high probabilities (\(0.98\), \(0.99\)), causing misdetection. In non-hallucinated cases, the key token \texttt{Cree} is correctly detected, while \texttt{"~"} is misidentified.

These results show that token-level detection is effective for key tokens but less reliable for frequent, less informative tokens. The ICR Probe primarily captures the residual stream pattern of key tokens, while irrelevant tokens act as noise, providing insights into its mechanism and guiding future optimizations.

\begin{table}[t]
\centering
\small
\begin{tabular}{p{7cm}}
\toprule
\textbf{Example}  \\
\midrule

\textbf{Question:} Tiendesitas is a part of an interim business district with an area of more than what? \\

\textbf{Response:} 18.5 hectares \\

\textbf{Token-level Detection Result:} \textbf{{\texttt{"1"(\textcolor{red}{0.34})}, \texttt{"8"(\textcolor{darkgreen}{0.99})}, \texttt{"."(\textcolor{red}{0.21})}, \texttt{"5"(\textcolor{red}{0.27})}, \texttt{"~hectares"(\textcolor{red}{0.04})}, \texttt{"~"(\textcolor{darkgreen}{0.96})}, \texttt{"\textbackslash n"(\textcolor{darkgreen}{0.98}) }}} \\

\textbf{Correct Answer:} 100 ha \\
\midrule

\textbf{Question:} What was the official name for the draft that Thomas E. O'Donnell was a public opponent of? \\

\textbf{Response:} The Vietnam War draft. \\

\textbf{Token-level Detection Result:} \textbf{\texttt{"The"(\textcolor{darkgreen}{0.98})}, \texttt{"~Vietnam"(\textcolor{red}{0.19})}, \texttt{"~War"(\textcolor{red}{0.28})}, \texttt{"~draft"(\textcolor{red}{0.17})}, \texttt{"."(\textcolor{darkgreen}{0.99})}, \texttt{"~"(\textcolor{red}{0.03})}, \texttt{"\textbackslash n"(\textcolor{darkgreen}{0.99})}} \\

\textbf{Correct Answer:} Conscription in the United States \\
\midrule

\textbf{Question:} Which First Nations group of over 200'000 members living in Canada signed an accord with the Quebec government, leading to the creation of Eeyou Istchee James Bay Territory? \\

\textbf{Response:} The Cree.  \\

\textbf{Token-level Detection Result:} \textbf{\texttt{"The"(\textcolor{darkgreen}{0.54})}, \texttt{"~Cree"(\textcolor{darkgreen}{0.55})}, \texttt{"."(\textcolor{darkgreen}{0.78})}, \texttt{"~"(\textcolor{red}{0.06})}, \texttt{"\textbackslash n"(\textcolor{darkgreen}{0.99})}} \\

\textbf{Correct Answer:} The Cree \\

\bottomrule
\end{tabular}
\caption{Token-level detection examples, with predicted probabilities displayed next to each token. Tokens are color-coded based on their probabilities: \textbf{Red} for hallucinations and \textbf{Green} for non-hallucinations.}
\label{tab:token_level_detection}
\end{table}

\section{Related Work} 

\paragraph{Hallucinations in LLMs}
Hallucinations are a prevalent phenomenon in large language models, typically defined as generated content that is nonsensical or deviates from the provided source content \cite{10.1145/3703155, 10.1145/3571730, li-etal-2023-halueval}. These hallucinations pose significant risks to the reliability and accuracy of LLMs. As a result, extensive research has been conducted to understand their underlying causes \cite{10.5555/3618408.3619699}, improve detection methods \cite{azariaInternalStateLLM2023}, and develop strategies for mitigating their effects \cite{chuang2024dola}.

\paragraph{Hallucination Detection via Hidden States} 
Several methods based on hidden states are proposed for hallucination detection. \citet{azariaInternalStateLLM2023} extract hidden states from specific layers and train probes, a method limited by its reliance on selected layers and tokens, which prevents it from capturing the dynamics across layers. \citet{ch-wang-etal-2024-androids} improve this approach by integrating probes, though this comes with high training costs and the need for extensive hyperparameter tuning. SEP \cite{Kossen2024SemanticEP} utilizes indirect signals, such as semantic entropy from multiple outputs, to train probes for hallucination detection.
Train-free methods are also introduced, such as calculating the regularized covariance matrix of hidden states \cite{chen2024INSIDELI} or the eigenvalues of attention kernels \cite{sriramananLLMCheckInvestigatingDetection2024a}. These methods eliminate the need for training, though they either require multiple generations or demonstrate limited detection performance. Additionally, \citet{sunReDeEPDetectingHallucination2024} propose the ReDeEP method, which focuses on RAG hallucinations by examining specific layers’ external context and parameter knowledge. While this approach captures the contribution of different modules, it lacks a comprehensive view of the entire residual stream and is dependent on hyperparameter selection.

\section{Conclusion}

In this work, we introduce a novel approach to hallucination detection in LLMs by focusing on the dynamic updates of hidden states rather than their static representations. The ICR Score quantifies the contribution of model components to the residual stream updates, providing a more precise measure for detection. By aggregating these scores across layers, the ICR Probe captures the cross-layer evolution of the residual stream, enabling more accurate and comprehensive detection. Extensive experiments demonstrate that the ICR Probe outperforms baseline methods, with strong generalization across datasets and enhanced interpretability. Ablation studies and case studies further confirm its effectiveness and provide valuable insights into the underlying mechanisms of the approach.

\section*{Limitations}

The proposed ICR Probe offers an effective approach to hallucination detection but has several limitations. First, it is applicable to open-source models, as it requires access to hidden states, which limits its use in proprietary models. Second, while this work focuses on hallucination detection, it does not directly address mitigating hallucinations in LLMs. We hope that future research, building on the insights and metrics proposed in this work, will explore interventions in the residual stream update process to reduce the tendency for hallucinations. Finally, although our experiments span diverse datasets (e.g., HaluEval, SQuAD), extending evaluations to more complex generation tasks would better validate the robustness of the approach.

\section*{Ethics Statement}
In this research, we prioritize user privacy and data protection, as our methodology does not rely on personal or sensitive data. All experiments were conducted using publicly accessible datasets, ensuring adherence to privacy and ethical guidelines. While the goal of this work is to enhance the reliability and performance of large language models, we recognize the potential risks of misuse, including the creation of misleading or harmful AI-generated content. We are committed to promoting ethical AI practices, addressing concerns regarding misuse, and ensuring that our research contributes positively to society. Additionally, AI assistants were used solely for text polishing and were not involved in any aspect of the core research process.

\section*{Acknowledgements}
This work was supported by Beijing Science and Technology Program (Z231100007423011) and Key Laboratory of Science, Technology and Standard in Press Industry (Key Laboratory of Intelligent Press Media Technology). We appreciate the anonymous reviewers for their helpful comments. Xiaojun Wan is the corresponding author.

\bibliography{main}

\appendix
\section{Calculation Details and Explanation of ICR Score}
\label{appendix:icr_score}
ICR Score is computed by Equation~\ref{eq:icore_score}, where JSD quantifies the similarity between two distributions, and possesses the property of being symmetric and bounded. 
\subsection{Introduction to JSD}
The Jensen-Shannon Divergence (JSD) \cite{61115} is calculated using the following formula:

\[
\text{JSD}(P || Q) = \frac{1}{2} \left( D_{\text{KL}}(P || M) + D_{\text{KL}}(Q || M) \right)
\]

where:

\begin{itemize}
    \item \( P \) and \( Q \) are two probability distributions.
    \item \( M = \frac{1}{2}(P + Q) \) is the average of the two distributions.
    \item \( D_{\text{KL}}(P || M) \) and \( D_{\text{KL}}(Q || M) \) are the Kullback-Leibler (KL) divergences between \( P \) and \( M \), and \( Q \) and \( M \), respectively.
\end{itemize}

The KL divergence is defined as:

\[
D_{\text{KL}}(P || Q) = \sum_{i} P(i) \log \left( \frac{P(i)}{Q(i)} \right)
\]

where \( P(i) \) and \( Q(i) \) represent the probabilities of the \(i\)-th event in the distributions \(P\) and \(Q\).

JSD quantifies the similarity between two probability distributions \( P \) and \( Q \). It is symmetric and produces a value between 0 and 1. A JSD of 0 indicates identical distributions, while a value close to 1 indicates significant difference. The average distribution \( M \) is used to ensure symmetry and handle differing supports between \( P \) and \( Q \).

\paragraph{Role and Impact of ICR Score}
The ICR Score quantifies the relative contribution of attention and FFN modules to the hidden state update in each layer. A small ICR score suggests the attention mechanism primarily drives the update, while a large score indicates the FFN plays a dominant role. By measuring these contributions, the ICR score helps reveal how attention and FFN modules influence the residual stream, offering insights into the model's internal dynamics and the distribution of responsibility between the two components.

\subsection{ICR Score and Probe Workflow}
We summarize the procedures for computing the ICR Score and deploying the ICR Probe; see Algorithm~\ref{alg:icr} and Algorithm~\ref{alg:icr-probe}, respectively.

\begin{algorithm*}[h]
\caption{ICR Score Computation}\label{alg:icr}
\begin{algorithmic}[1]
\Require Hidden states $\{x_i^{\ell-1},x_i^\ell\}_{i=1}^N$ for $\ell=1,\dots,L$, 
         attention logits $\{\mathrm{Attn}_i^{\ell,h}\}_{h=1}^H$, top-$k$
\Ensure  ICR scores $\{\mathrm{ICR}_i^\ell\}$ for all tokens and layers
\For{$\ell=1$ \textbf{to} $L$}
  \For{$i=1$ \textbf{to} $N$}
    \State $a_i^\ell \gets \mathrm{softmax}\!\Bigl(\frac{1}{H}\sum_{h=1}^H \mathrm{Attn}_i^{\ell,h}\Bigr)$
           \Comment{\textcolor{gray}{normalize multi-head attention}}
    \State $\Delta x_i^\ell \gets x_i^\ell - x_i^{\ell-1}$ 
           \Comment{\textcolor{gray}{token update vector}}
    \For{$j=1$ \textbf{to} $N$}
       \State $p_{i,j}^\ell \gets \dfrac{(\Delta x_i^\ell)^\top x_j^\ell}{\|x_j^\ell\|}$
    \EndFor
    \State $\mathrm{proj}_i^\ell \gets \mathrm{softmax}\bigl(\{p_{i,j}^\ell\}_{j=1}^N\bigr)$
           \Comment{\textcolor{gray}{projection distribution}}
    \State $S \gets \text{TopIndices}(a_i^\ell, k)$ 
           \Comment{\textcolor{gray}{indices of top-$k$ attention weights}}
    \State $\mathrm{ICR}_i^\ell \gets 
           \mathrm{JSD}\bigl(\mathrm{proj}_i^\ell[S]\,\Vert\,a_i^\ell[S]\bigr)$
  \EndFor
\EndFor
\end{algorithmic}
\end{algorithm*}

\begin{algorithm*}[t]
\caption{ICR Probe: Training and Inference}\label{alg:icr-probe}
\begin{algorithmic}[1]
\Require 
  LLM with $L$ transformer layers, training corpus $\mathcal D=\{(q,a,y)\}$ with question $q$, generated answer $a$, label $y\in\{0,1\}$ (\textit{0: faithful, 1: hallucinated});  
  top-$k$ for ICR Score; learning rate $\eta$
\Ensure 
  Trained probe parameters $\theta$ and inference routine
\Procedure{Train}{$\mathcal D$}
  \State Initialize probe weights $\theta$ (logistic regression)
  \ForAll{$(q,a,y)\in\mathcal D$}
      \State Run LLM once on $(q,a)$, cache hidden states $\{x_i^\ell\}$ and attentions $\{\text{Attn}_i^{\ell,h}\}$
      \ForAll{answer tokens $i$ \textbf{and} layers $\ell=1..L$}
          \State Compute $\text{ICR}_i^\ell$ by Algorithm~\ref{alg:icr}  \Comment{\textcolor{gray}{single forward‐pass}}
      \EndFor
      \State $\displaystyle f \gets \Bigl[\;\underset{i}{\text{mean}}\bigl(\text{ICR}_i^\ell\bigr)\Bigr]_{\ell=1}^L$   \Comment{\textcolor{gray}{layer-wise mean pool}}
      \State $\hat{y}\gets\sigma(\theta^\top f)$ \Comment{\textcolor{gray}{sigmoid}}
      \State $\theta\gets\theta-\eta\,(\hat{y}-y)\,f$  \Comment{\textcolor{gray}{SGD update}}
  \EndFor
  \State \textbf{return} $\theta$
\EndProcedure
\vspace{0.5em}
\Procedure{Infer}{$q,a,\theta$}
  \State Run LLM on $(q,a)$, cache $\{x_i^\ell\},\{\text{Attn}_i^{\ell,h}\}$
  \ForAll{answer tokens $i$ \textbf{and} layers $\ell$}
      \State $\text{ICR}_i^\ell\gets$ Algorithm~\ref{alg:icr}
  \EndFor
  \State $f\gets[\text{mean}_i(\text{ICR}_i^\ell)]_{\ell=1}^L$
  \State $\hat{y}\gets\sigma(\theta^\top f)$
  \State \textbf{return} $\hat{y}$  \Comment{\textcolor{gray}{hallucination probability}}
\EndProcedure
\end{algorithmic}
\end{algorithm*}

\section{Details about Experiment}
\paragraph{Computing Infrastructure} Our experiments were conducted on a server equipped with 10 NVIDIA GeForce RTX 3090 GPUs (24 GB memory each), running CUDA 12.0 and Ubuntu 20.04.5 LTS. Across multiple rounds of experiments, the total computational budget amounted to 600-800 GPU hours.

\subsection{Details about Datasets}
\label{appendix:dataset}
For our experiments, we randomly sampled 10,000 instances from each dataset. Specifically, we used the QA subset of the HaluEval dataset and the rc.nocontext subset of TriviaQA. The datasets used in this study are publicly available and adhere to their respective licenses, which allow for academic research and non-commercial use.
Each dataset is split into 80\%-20\% for training and testing. We train and test each dataset, reporting the corresponding results. \textbf{This experimental setup for baseline methods matches ours exactly.}

\subsection{Baselines Methods}
\label{appendix:baselines}
This section provides a brief overview of several baseline methods for hallucination detection, including metrics like PPL and LN-Entropy, as well as techniques such as LLM-check, SAPLMA, and SEP.

\paragraph{PPL:}
 PPL assesses the likelihood of hallucinations in LLM-generated responses by calculating perplexity. A higher perplexity value reflects increased uncertainty in the model's output, as hallucinations are often associated with the model's lack of confidence or inaccurate knowledge. 

\paragraph{Length-Normalized Entropy (LN-Entropy):}
 LN-Entropy is designed to quantify sequence-level uncertainty across multiple generations. This metric normalizes entropy concerning sequence length, facilitating a more precise assessment of the stability and reliability of generated content. A high value of the normalized entropy may serve as an indicator of the presence of hallucinations or other forms of unreliable output.

\paragraph{LLM-Check:}
 LLM-check uses the method of attention mechanism kernel similarity analysis to conduct hallucination detection. They found that in large language models, for an input sequence of length $m$ with $a$ self-attention heads per layer, the kernel similarity map $Ker_{i}$ of each head is a lower-triangular $(m × m)$ matrix. By calculating $log det(Ker_{i})=\sum_{j = 1}^{m} log Ker_{i}^{jj}$ and aggregating the mean log - determinants of all heads to obtain the "Attention Score", it can effectively capture the characteristics related to hallucinations. This method is computationally efficient and performs well in hallucination detection across different datasets and settings. In black-box settings, an auxiliary LLM with teacher-forcing is used to acquire the scores for detection. 

\paragraph{SAPLMA:}
 SAPLMA trains a classifier to detect hallucinations using the hidden states of specific layers in LLMs. It captures internal signals from these layers, which the classifier uses to identify when the model is likely to generate hallucinated content.

\paragraph{SEP (Semantic Entropy Probe):}
 SEP leverages linear probes trained on the hidden states of large language models. It detects hallucinations by analyzing the semantic entropy of tokens before generation. This approach assumes that the semantic entropy of tokens can provide insights into whether the subsequently generated content is hallucinatory.

\subsection{Ablation Study on the Number of Selected Tokens ($k$)}
To validate our choice of $k=20$ when computing the ICR Score, we conduct an ablation study on the Gemma-2 model across four benchmark datasets. Table~\ref{tab:ablation-k} summarizes detection performance for $k\in\{5,20,30,\text{ALL}\}$.  

\begin{table}[h]
  \centering
  \small
  \caption{Detection performance (higher is better) under different values of $k$.}
  \label{tab:ablation-k}
  \begin{tabular}{c|cccc}
    \toprule
    $k$   & HaluEval & SQuAD  & HotpotQA & TriviaQA \\
    \midrule
    5     & 0.8117   & 0.7830 & 0.8172   & 0.7926   \\
    20    & \textbf{0.8436}   & \textbf{0.8142} & \textbf{0.8409}   & 0.8001   \\
    30    & 0.8431   & 0.8005 & 0.8360   & \textbf{0.8005}   \\
    ALL   & 0.8289   & 0.7853 & 0.8327   & 0.7918   \\
    \bottomrule
  \end{tabular}
\end{table}

\paragraph{Why not use all tokens?}  
Using only the top-$k$ tokens yields consistently better detection accuracy than aggregating over all tokens. For example, on HaluEval the ICR Score with $k=20$ outperforms the “ALL” baseline by +1.47\%. Intuitively, selecting the highest-attention tokens filters out noisy or irrelevant positions in the attention map, leading to more robust attribution.

\paragraph{Sensitivity to $k$.}  
The performance peak at $k=20$ is stable: increasing to $k=30$ changes the results by less than 0.5\% across all datasets. Conversely, too small a $k$ (e.g.\ $k=5$) under-samples the important tokens and degrades detection accuracy by over 3\% on some benchmarks.

Based on these results, we fix $k=20$ for all further experiments in Table~1. Full ablation across other LLMs will be provided in the revised Appendix due to space constraints.

\subsection{ICR Probe Training}
\label{appendix:probe}
\paragraph{Input and Output}

The ICR Probe is a multi-layer perceptron (MLP) classifier trained to predict hallucinations using the pooled ICR score. Given a model generation, the ICR matrix \( \mathbb{R}^{N \times L} \) is token-wise pooled to obtain a \( 1 \times L \) vector, which serves as the input to the probe. The label for each instance is derived from annotated hallucination datasets.

\paragraph{Model Architecture }
The probe consists of four fully connected layers, with batch normalization and dropout (\(p = 0.3\)) applied after each hidden layer. To strike a balance between performance and efficiency, we conducted preliminary experiments to select an appropriate probe structure. The results are shown in Table \ref{tab:architecture_select}, taking Gemma - 2 and TriviaQA as examples. When the probe has four hidden layers, an optimal balance between parameter efficiency and performance is achieved.

\begin{table}[t]
\centering
\small
\setlength{\tabcolsep}{2pt} 
\begin{tabular}{@{}c|ccccc@{}}
\toprule
\textbf{Hidden layers} & \textbf{1} & \textbf{2} & \textbf{3} & \textbf{4} & \textbf{5} \\ \midrule
\textbf{Performance} & 0.6410 & 0.7398 & 0.7889 & 0.8000 & 0.8006 \\ \bottomrule
\end{tabular}
\caption{Comparison results of performance of different probe structures.}
\label{tab:architecture_select}
\end{table}

\begin{figure*}[!t]
\centering 
\begin{subfigure}[b]{0.46\textwidth}
    \centering
    \includegraphics[width=\textwidth]{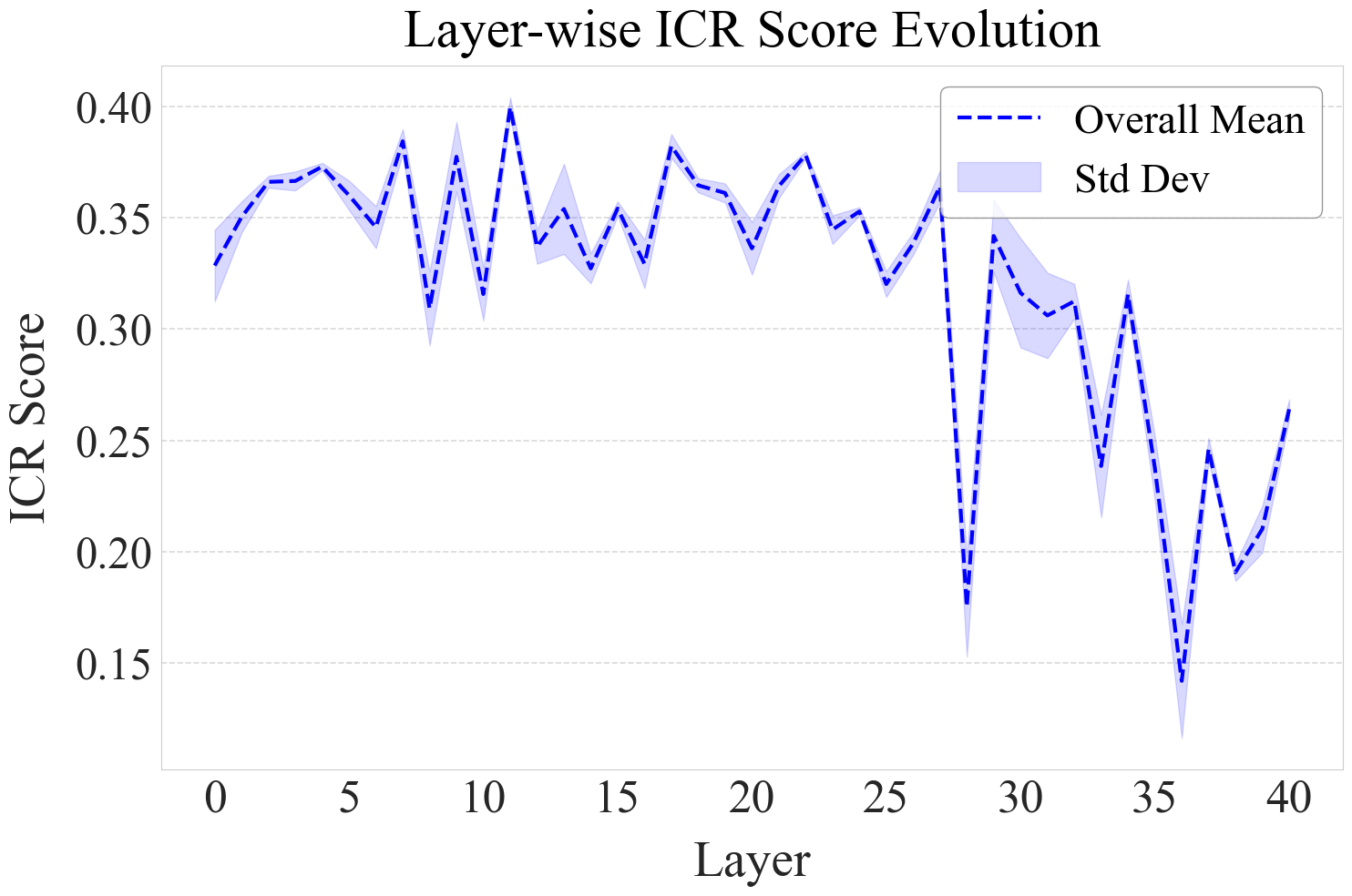}
    \caption{Gemma-2: Layer-wise AUROC for hallucination detection \textbf{using direct ICR scores}, peaking at layer \textbf{21}.}
\end{subfigure}
\hspace{5mm} 
\begin{subfigure}[b]{0.46\textwidth}
    \centering
    \includegraphics[width=\textwidth]{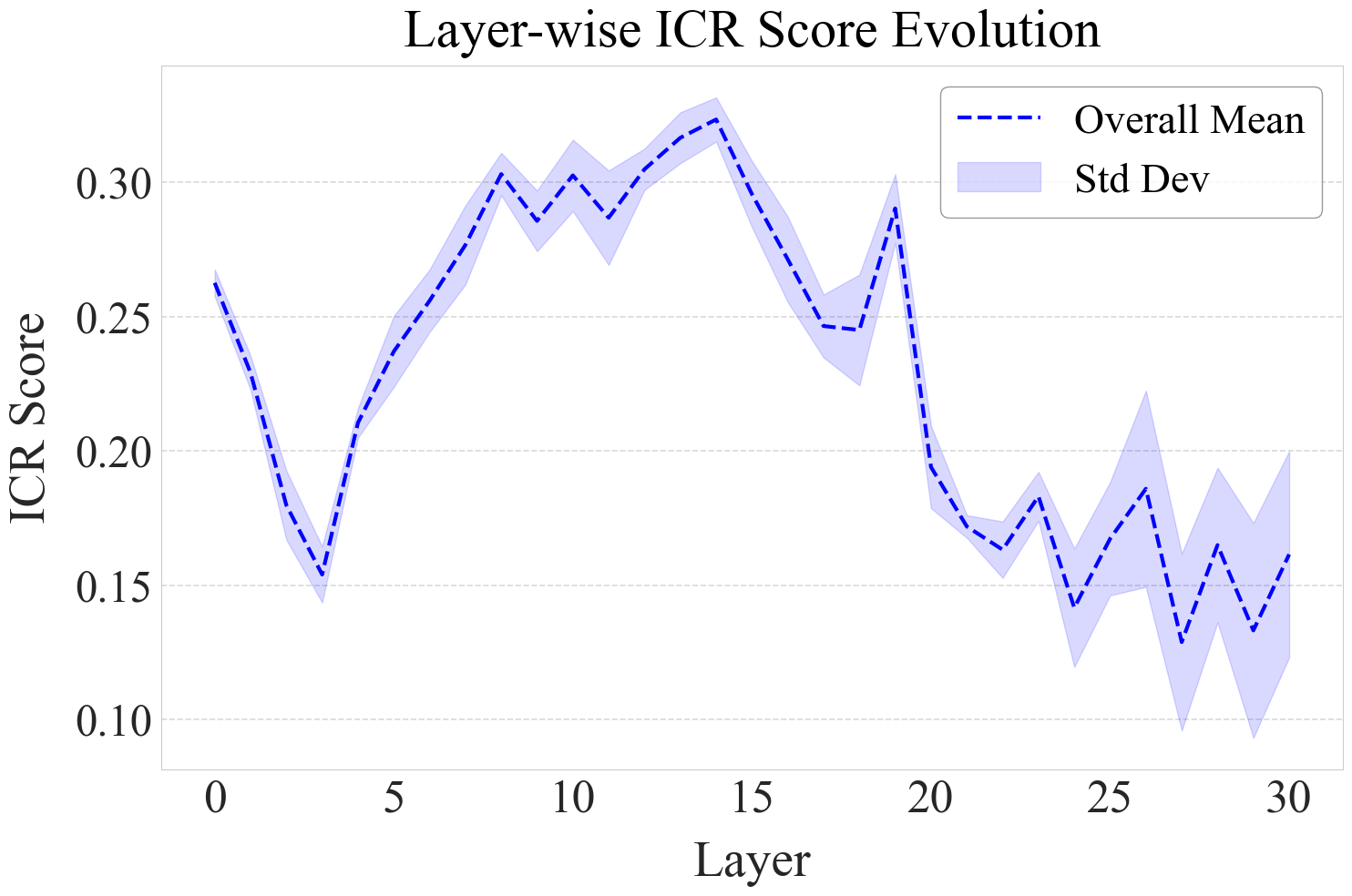}
    \caption{Llama-3: Layer-wise AUROC for hallucination detection \textbf{using direct ICR scores}, peaking at layer \textbf{10}.}
\end{subfigure}
\caption{Layer-wise mean ICR scores.}
\label{fig:icr_score_appendix}
\end{figure*}

\begin{figure*}[!t]
\centering 
\begin{subfigure}[b]{0.46\textwidth}
    \centering
    \includegraphics[width=\textwidth]{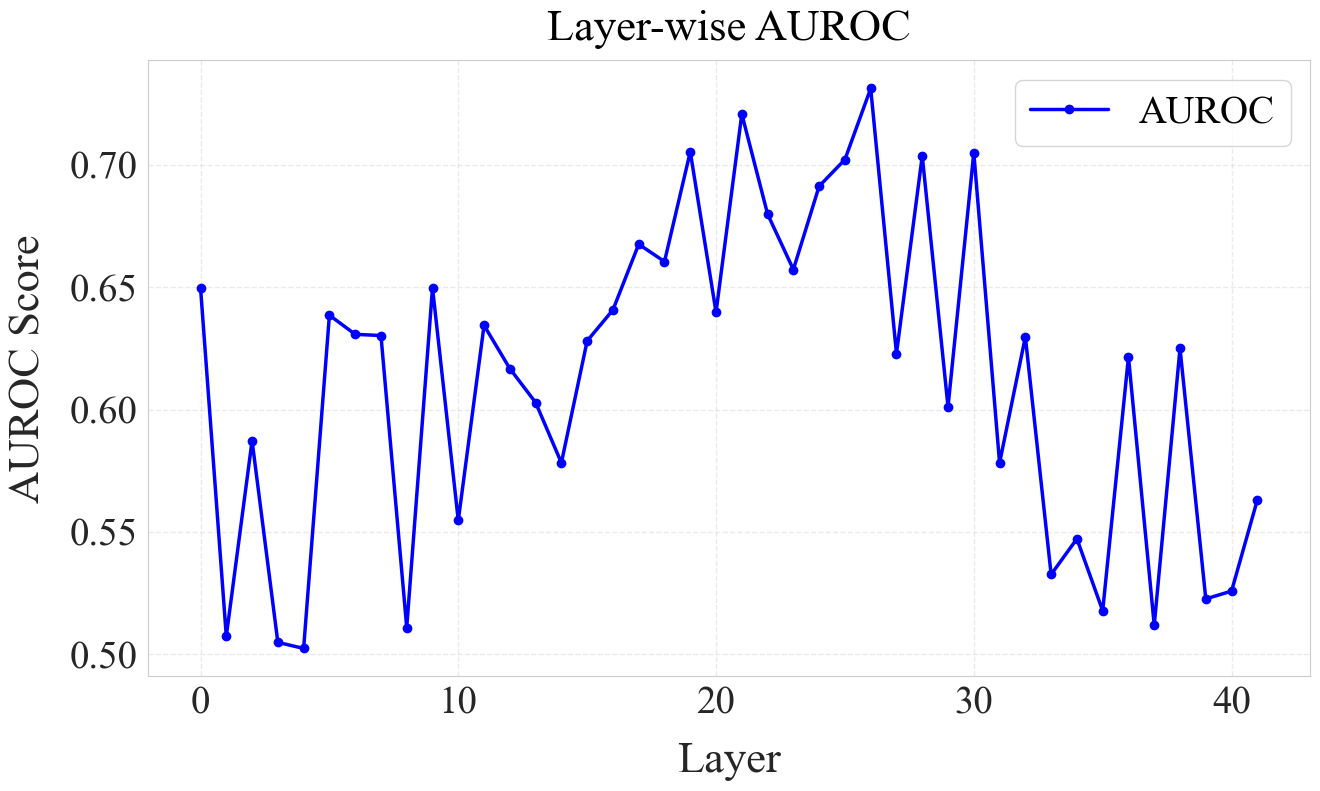}
    \caption{Gemma-2: Layer-wise AUROC for hallucination detection \textbf{using direct ICR scores}, peaking at layer \textbf{21}.}
\end{subfigure}
\hspace{5mm} 
\begin{subfigure}[b]{0.46\textwidth}
    \centering
    \includegraphics[width=\textwidth]{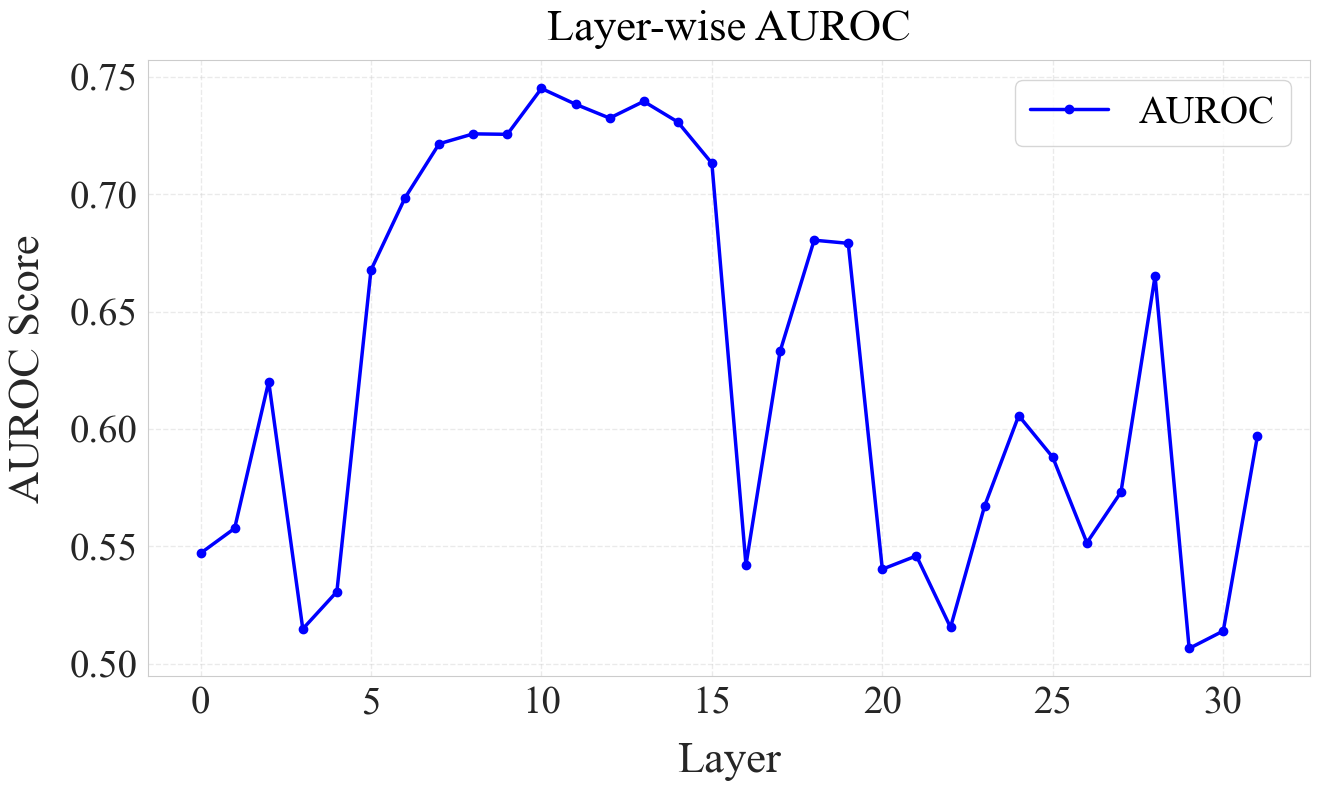}
    \caption{Llama-3: Layer-wise AUROC for hallucination detection \textbf{using direct ICR scores}, peaking at layer \textbf{10}.}
\end{subfigure}
\caption{Layer-wise AUROC.}
\label{fig:layer_auroc_appendix}
\end{figure*}

\begin{table*}[t]
  \centering
  \begin{tabular}{lcccc}
    \toprule
    \textbf{LLM}    & \textbf{HaluEval} & \textbf{SQuAD} & \textbf{HotpotQA} & \textbf{TriviaQA} \\
    \midrule
    Gemma-2         & 0.7850            & 0.7321         & 0.7465            & 0.7883            \\
    Qwen2.5         & 0.7592            & 0.6903         & 0.7345            & 0.7582            \\
    Llama-3         & 0.7458            & 0.7331         & 0.7340            & 0.7206            \\
    \bottomrule
  \end{tabular}
  \caption{AUROC of the Semantic Entropy baseline.}
  \label{tab:semantic-entropy}
\end{table*}

The architecture is as follows:
\[
L \rightarrow 128 \rightarrow 64 \rightarrow 32 \rightarrow 1
\]
Each hidden layer employs the \textbf{Leaky ReLU} activation function (\(\alpha = 0.01\)), and the final output is passed through a \textbf{sigmoid} activation to obtain a probability score.

\paragraph{Details of Training Probe}
 
 \begin{itemize}
     \item The model is trained using the binary cross-entropy loss with the Adam optimizer (\(\text{lr} = 5 \times 10^{-4}\)). We employ learning rate scheduling using \texttt{ReduceLROnPlateau}), where the learning rate is reduced by a factor of 0.5 if the validation loss does not improve for 5 consecutive epochs.

     \item We use Kaiming initialization for linear layers and set batch normalization parameters to 1 (scale) and 0 (bias).

     \item Dropout (\(p=0.3\)) is applied after each hidden layer to prevent overfitting.

     \item  The model is trained for 50 epochs with a batch size of 32. 

     \item To obtain stable and reliable results, we perform multiple runs and take the average.
 \end{itemize}

\begin{table*}[t]
  \centering

  \begin{tabular}{l l cccc}
    \toprule
    \textbf{Model} & \textbf{Method} & \textbf{HaluEval} & \textbf{SQuAD} & \textbf{HotpotQA} & \textbf{TriviaQA}\\
    \midrule
    \multirow{3}{*}{Qwen2.5--14B} 
      & SAPLMA               & 0.7720 & 0.6984 & 0.7214 & 0.7759 \\
      & SEP                  & 0.7016 & 0.6653 & 0.6874 & 0.7103 \\
      & \textbf{ICR Probe}   & \textbf{0.8021} & \textbf{0.7632} & \textbf{0.7751} & \textbf{0.7846} \\
    \midrule
    \multirow{3}{*}{Qwen2.5--3B} 
      & SAPLMA               & 0.7538 & 0.6915 & 0.7747 & 0.7523 \\
      & SEP                  & 0.6689 & 0.6560 & 0.6729 & 0.6945 \\
      & \textbf{ICR Probe}   & \textbf{0.7917} & \textbf{0.7784} & \textbf{0.7905} & \textbf{0.7631} \\
    \bottomrule
  \end{tabular}
    \caption{Performance of the ICR Probe on Qwen2.5 models of different scales.}
  \label{tab:scale}
\end{table*}

\subsection{Ablation Study Settings}
\label{appendix:ablation}

In this appendix, we provide details on the three experimental settings used in the ablation study, including the calculation of ICR Score in each setting and the corresponding meanings. 

\paragraph{1. NONE Setting}
In the \textbf{NONE} setting, both the hidden state (\(\text{Proj}_i^\ell\)) and attention (\(\text{Attn}_i^\ell\)) signals are excluded. This results in a random performance scenario, where the AUROC is equal to 0.5, indicating that the model's performance is no better than random guessing. In this setting, the ICR Score is not computed because no useful information is provided by either the hidden states or attention scores. This setup serves as a baseline to assess the importance of both components in hallucination detection.

\paragraph{2. HS ONLY Setting}
In the \textbf{HS ONLY} setting, only the hidden state information (\(\text{Proj}_i^\ell\)) is used, while the attention scores (\(\text{Attn}_i^\ell\)) are set to a uniform distribution. This effectively reduces the ICR Score to the entropy of the hidden state update directions. Specifically, the ICR Score in this setting is calculated as:

\[
\text{ICR}_{i,\text{HS ONLY}}^\ell = \text{JSD}(\text{Proj}_i^\ell, \mathbf{p}_{\text{uniform}})
\]

where \( \mathbf{p}_{\text{uniform}} \) is a vector representing a uniform distribution, and JSD stands for the Jensen-Shannon Divergence.

JSD measures the similarity between two distributions. When we compare the hidden state updates (\(\text{Proj}_i^\ell\)) to a uniform distribution, it is equivalent to measuring the entropy of the hidden state updates. This is because the entropy represents the amount of uncertainty or disorder within a distribution, and the uniform distribution can be considered as the maximum entropy distribution in the absence of any other information. Therefore, the JSD between the hidden state updates and a uniform distribution is a measure of how “disordered” or “uncertain” the update directions are, and this is equivalent to computing the entropy of \(\text{Proj}_i^\ell\).

\paragraph{3. HS + ATTN Setting}
In the \textbf{HS + ATTN} setting, both the hidden state (\(\text{Proj}_i^\ell\)) and attention signals (\(\text{Attn}_i^\ell\)) are integrated to compute the full ICR Score. The calculation of ICR Score in this setting is:

\[
\text{ICR}_{i,\text{HS + ATTN}}^\ell = \text{JSD}(\text{Proj}_i^\ell, \text{Attn}_i^\ell)
\]

\paragraph{Summary of Results}
The results from these three settings show that the hidden state alone (\(\text{Proj}_i^\ell\)) is already sufficient for effective hallucination detection, as evidenced by the AUROC being significantly above 0.5. However, integrating both hidden state and attention signals leads to further improvement in detection performance, highlighting the complementary nature of these components. The entropy of the hidden state update directions captures meaningful information about the model's updates, while the inclusion of attention signals helps refine this information.

\section{Additional Experimental Results}

\subsection{Empirical Study Results}

In this section, we provide further empirical analysis supporting the findings and conclusions presented in Section \ref{sec:empirical_study}. Figure \ref{fig:icr_score_appendix} illustrates the mean ICR scores, accompanied by standard deviation bands, across four datasets for both Llama-3 and Gemma-2. Narrow standard deviation bands across all models indicate the stability and consistency of the ICR scores across datasets.

Figure \ref{fig:layer_auroc_appendix} presents the layer-wise AUROC for hallucination detection using direct ICR scores on Llama-3 and Gemma-2. Notably, the highest AUROC on Llama-3 occurs at layer 10 (0.7451), while on Gemma-2, the highest AUROC is observed at layer 21 (0.7313). These results further validate the discriminative power of the ICR approach for hallucination detection.

\subsection{Additional Baseline}
\label{app:baseline}

To provide a broader context, we report the cross‐dataset AUROC of the Semantic Entropy baseline~\cite{kuhn2023semantic} alongside our ICR Probe. Note that Semantic Entropy relies on multiple generations and semantic clustering, whereas ICR Probe requires only a single forward pass, making the comparison computationally imbalanced. Table~\ref{tab:semantic-entropy} summarizes the Semantic Entropy performance.

\paragraph{Discussion}
We deliberately exclude entailment-based methods (e.g., AlignScore, LLM prompting \cite{zha2023alignscore}) because they operate under a fundamentally different paradigm. Entailment‐based approaches require an explicit premise-hypothesis setup, verifying consistency between a reference text and generated output, whereas our hallucination detection operates directly on hidden state representations without external references, enabling real-time single-pass inference. Consequently, entailment methods cannot be applied in scenarios lacking a well-defined premise (as in many open-ended QA or generation tasks; see Table~\ref{tab:token_level_detection}), making them structurally incompatible with our evaluation protocol and task objectives.

\begin{table*}[t]
\centering
\begin{tabular}{@{}cc|cccc@{}}
\toprule
\multicolumn{2}{c|}{\multirow{2}{*}{\textbf{Qwen2.5}}} & \multicolumn{4}{c}{\textbf{Test}} \\
\multicolumn{2}{c|}{} & HaluEval & SQuAD & HotpotQA & TriviaQA \\ \midrule
\multirow{4}{*}{\textbf{Train}} & HaluEval & \textbf{0.8003} & 0.6835 & 0.7589 & 0.6626 \\
 & SQuAD & 0.7360 & \textbf{0.7456} & 0.7414 & 0.6900 \\
 & HotpotQA & 0.7822 & 0.7028 & \textbf{0.7917} & 0.6387 \\
 & TriviaQA & 0.7576 & 0.6962 & 0.7222 & \textbf{0.7684} \\ \bottomrule
\end{tabular}
\caption{Cross-dataset generalization evaluation for Qwen2.5. Each cell shows the AUROC when the probe is trained on the row dataset and evaluated on the column dataset.}
\label{tab:gen_qwen_appendix}
\end{table*}

\begin{table*}[t]
\centering
\begin{tabular}{@{}cc|cccc@{}}
\toprule
\multicolumn{2}{c|}{\multirow{2}{*}{\textbf{Llama-3}}} & \multicolumn{4}{c}{\textbf{Test}} \\
\multicolumn{2}{c|}{} & HaluEval & SQuAD & HotpotQA & TriviaQA \\ \midrule
\multirow{4}{*}{\textbf{Train}} & HaluEval & \textbf{0.7603} & 0.7210 & 0.7547 & 0.6289 \\
 & SQuAD & 0.7342 & \textbf{0.7634} & 0.7294 & 0.5658 \\
 & HotpotQA & 0.7775 & 0.7347 & \textbf{0.7982} & 0.5933 \\
 & TriviaQA & 0.6972 & 0.6129 & 0.6749 & \textbf{0.7325} \\ \bottomrule
\end{tabular}
\caption{Cross-dataset generalization evaluation for Llama-3. Each cell shows the AUROC when the probe is trained on the row dataset and evaluated on the column dataset.}
\label{tab:gen_llama3_appendix}
\end{table*}

\subsection{Performance Across Model Scales}\label{app:model-scale}

To evaluate the ICR Probe across different model sizes, we include both a smaller (3B) and a larger (14B) variant of Qwen2.5. Table~\ref{tab:scale} reports the AUROC of ICR Probe alongside the strongest hidden-state baselines, SAPLMA and SEP.

\paragraph{Discussion.}
Across both additional scales, \textsc{ICR Probe} remains the top performer on every dataset.  
Compared with \textsc{SAPLMA}, it gains on average \textbf{3.9\,\%} AUROC at 14B and \textbf{3.8\,\%} at 3B; against \textsc{SEP} the average margin widens to \textbf{9.0\,\%} and \textbf{10.8\,\%}, respectively.
These results confirm that the probe’s layer-wise residual-stream cues are largely scale-invariant, supporting its applicability to a broad spectrum of model sizes.


\subsection{Generalization}

In Section~\ref{generalization}, we demonstrate the generalization ability of the ICR Probe on Gemma-2. Here, we provide additional generalization results. Tables~\ref{tab:gen_llama3_appendix} and~\ref{tab:gen_qwen_appendix} present the cross-dataset generalization evaluation results for Llama-3 and Gemma-2, respectively. The AUROC values for both models exceed 0.6, highlighting the strong generalization capability of our approach.

\end{document}